\documentclass[10pt,journal,compsoc]{IEEEtran}
\usepackage{amsfonts}
\usepackage{booktabs} 
\usepackage{multirow}
\usepackage{array}
\usepackage{xcolor}
\definecolor{CQColor}{rgb}{0.0,0.0,1.0} 

\definecolor{CQRColor}{rgb}{1.0,0.0,1.0} 

\usepackage{ragged2e}
\renewcommand{\raggedright}{\leftskip=0pt \rightskip=0pt plus 0cm}

\ifCLASSOPTIONcompsoc
  \usepackage[nocompress]{cite}
\else
  \usepackage{cite}
\fi
\ifCLASSINFOpdf
  \usepackage[pdftex]{graphicx}
\else
\fi
\usepackage{amsmath}
\usepackage{array}

\ifCLASSOPTIONcompsoc
 \usepackage[caption=false,font=footnotesize,labelfont=sf,textfont=sf]{subfig}
\else
  \usepackage[caption=false,font=footnotesize]{subfig}
\fi
\begin{document}
%
\title{Deep Multi-View Enhancement Hashing for\\ Image Retrieval}
%
%
%
%

%

\author{Chenggang~Yan, 
		Biao~Gong,
		Yuxuan~Wei,
		and~Yue~Gao,~\IEEEmembership{Senior~Member,~IEEE}
	
\IEEEcompsocitemizethanks{
\IEEEcompsocthanksitem Chenggang Yan, Biao Gong, Yuxuan Wei, and Yue Gao. 

E-mail: cgyan@hdu.edu.cn; a.biao.gong@gmail.com; weiyuxua19@mails. tsinghua.edu.cn; kevin.gaoy@gmail.com (Corresponding author: Y. Gao.)
}
}

\IEEEtitleabstractindextext{
\raggedright{
\begin{abstract}
Hashing is an efficient method for nearest neighbor search in large-scale data space by embedding high-dimensional feature descriptors into a similarity preserving Hamming space with a low dimension. However, large-scale high-speed retrieval through binary code has a certain degree of reduction in retrieval accuracy compared to traditional retrieval methods. We have noticed that multi-view methods can well preserve the diverse characteristics of data. Therefore, we try to introduce the multi-view deep neural network into the hash learning field, and design an efficient and innovative retrieval model, which has achieved a significant improvement in retrieval performance. In this paper, we propose a supervised multi-view hash model which can enhance the multi-view information through neural networks. This is a completely new hash learning method that combines multi-view and deep learning methods. The proposed method utilizes an effective view stability evaluation method to actively explore the relationship among views, which will affect the optimization direction of the entire network. We have also designed a variety of multi-data fusion methods in the Hamming space to preserve the advantages of both convolution and multi-view. In order to avoid excessive computing resources on the enhancement procedure during retrieval, we set up a separate structure called memory network which participates in training together. The proposed method is systematically evaluated on the CIFAR-10, NUS-WIDE and MS-COCO datasets, and the results show that our method significantly outperforms the state-of-the-art single-view and multi-view hashing methods.
\end{abstract}
}

}

\maketitle

\IEEEdisplaynontitleabstractindextext

%
\IEEEpeerreviewmaketitle

\IEEEraisesectionheading{\section{Introduction}\label{sec:introduction}}

%
%
%
%

\IEEEPARstart{W}{ith} the explosive growth of image data, the efficient large-scale image retrieval algorithms such as approximate nearest neighbour (ANN) search \cite{arya1998optimal,gong2011iterative,zhang2014supervised} which balances the retrieval time-consuming and retrieval efficiency on the large-scale dataset attract increasing attentions. In the field of large-scale image retrieval, learning-to-hash \cite{7915742} is a kind of emerging and highly efficient ANN search approach which aims to automatically learn optimal hash functions and generate image hash codes. The nearest neighbor is obtained by calculating the Hamming distance of these hash codes.

\begin{figure}[ht]
	\centering
	\includegraphics[scale=0.45]{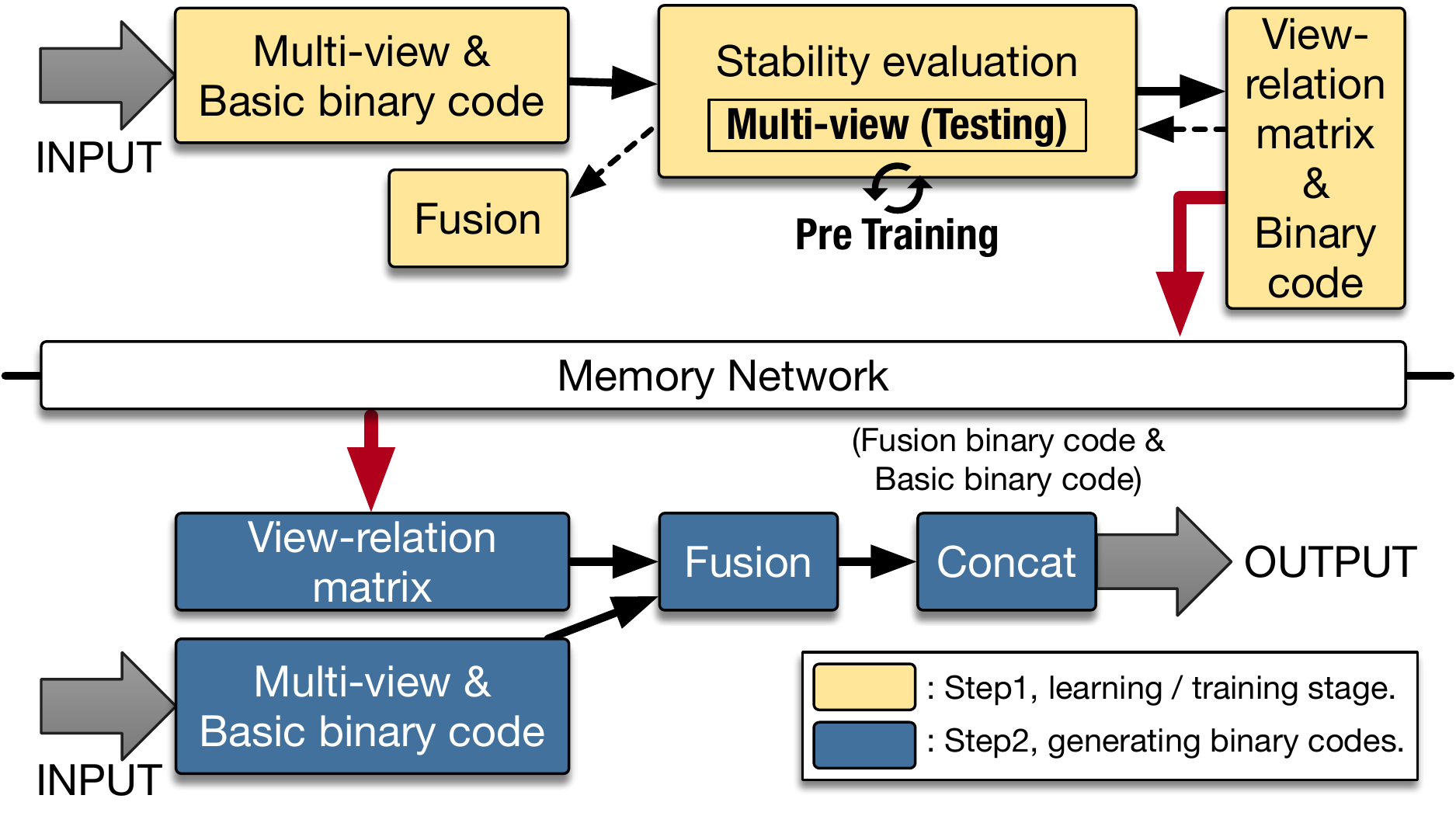}
	\caption{Illustration of the key idea of D-MVE-Hash. The core part of D-MVE-Hash called stability evaluation is used to obtain view-relation matrix in step1. Dotted arrows indicate the path of back propagation. The memory network is used to memorize the view-relation matrix which is calculated by the stability evaluation method during training. In hash codes generation step (e.g., step2), we directly use the memory network to produce the view-relation matrix. After feeding the generated multi-view and basic binary codes with the view-relation matrix to following fusion and concat layers, D-MVE-Hash outputs fusion binary representation of images for retrieval.}
	\label{graph:mmrnet}
\end{figure}

In recent years, several convolution artificial neural network hash methods \cite{liu2012supervised,zhen2016spectral,weiss2009spectral,cao2017hashnet,cao2018deep} which show significant improvement in the image retrieval task have been proposed. CNNH \cite{xia2014supervised} is a typical representative of these methods. Parallel to that, some impressive researches, such as \cite{lai2015simultaneous} propoesd by Lai \textit{et al.} which learns a well image representation tailored to hashing as well as a set of hash functions, highly improve the retrieval precision and let the efficient large-scale image retrieval enter the next stage.

However, these methods only focus on learning optimal hash functions from single-view data (i.e., obtaining image features from a single convolution feature space). Recently, a number of multi-view hashing methods \cite{shen2015multi,liu2015multiview,zhang2017semi,xie2017dynamic} which mainly depend on spectral, graph and deep learning techniques to achieve data structure preserving encoding have been proposed for efficient similarity search.
In a general sense, the hash method with multiple global features is so-called multi-view hashing. Other than this, the multi-view is also well known as multiple angles cameras of 3D model \cite{feng2018gvcnn}. For heterogeneous multimedia features such as multi-feature hashing of video data with local and global visual features \cite{song2013effective,8744407} and other orthogonal or associated features, researchers use multi/cross-modal hashing to solve the complexity of the fusion problem of multiple modalities. It is disparate from the multi-view hashing. In this paper, we define the multi-view as multi-angle representations in different visual feature spaces.
The supplement of additional information is one of the most prominent contribution for these multi-view methods.
Recently, an increasing number of researches explicitly or implicitly capture the relations among different views \cite{shen2016semi,jia2019deep} and modalities \cite{hu2018collective,xu2017learning} to enhance the multi/cross-view/modal approaches.
Compared to these methods, the view-relation matrix we proposed can directly be employed in original feature view spaces. It avoids information missing during the process of subspace mapping resulting in a better capture of relations.
Furthermore, our approach optimizes the feature extraction network by both objective function and view relation calculation process to obtain the best multi-view features which are used to calculate the matrix.

Our motivation to design deep multi-view enhancement hashing (D-MVE-Hash) arises from two aspects: (1) By splicing and normalizing the fluctuations of different image features under multiple feature views, we obtain the volatility matrix which regulates the view relevance based on fluctuation strength to produce a quality view-relation matrix. The view-independent and robustness are the special properties of this view-relation matrix. We have conducted detailed ablation experiments in Sec.~\ref{sec:exper3} to support this conclusion.
(2) Since typical multi-view hash image retrieval algorithm still needs to manually extract image features \cite{xie2017dynamic}, which makes the multi-view information and the deep network less impactful in learning-to-hash and image retrieval. For this reason, we want to design an end-to-end architecture with data fusion methods which are carried out in the Hamming space.

\begin{figure*}
	\centering
	\includegraphics[scale=0.6]{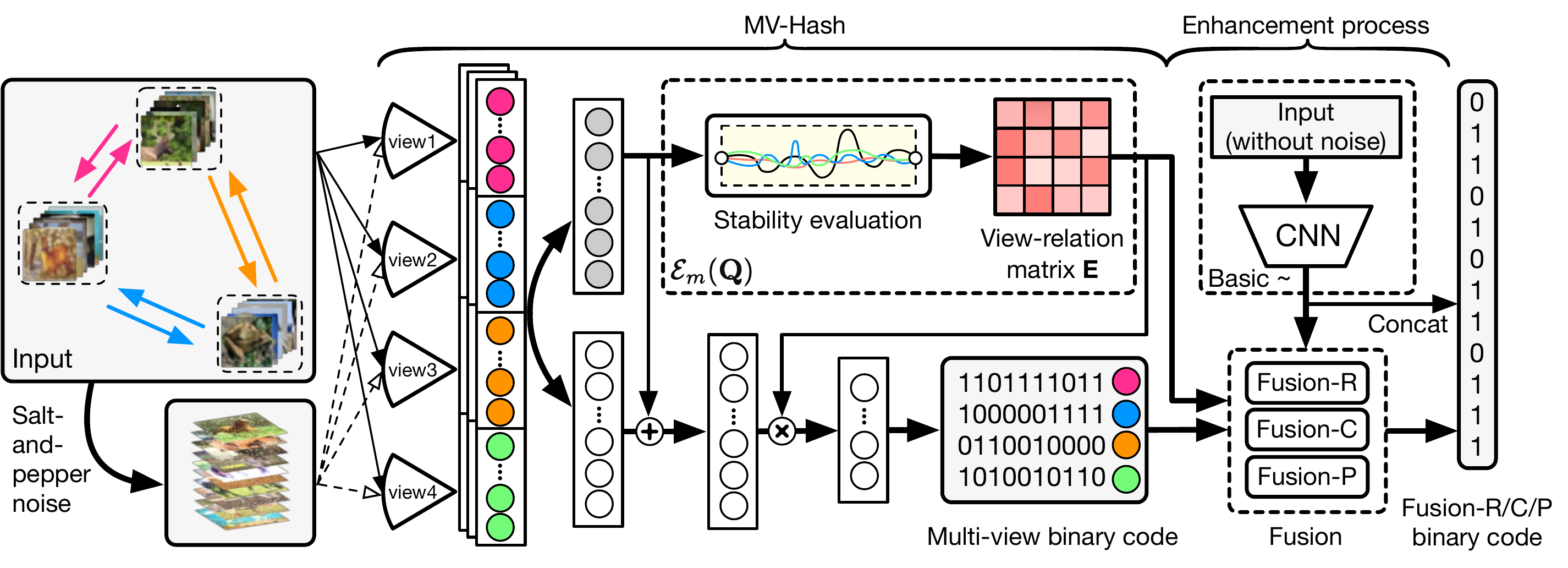}
	\caption{The proposed deep multi-view enhancement hashing (D-MVE-Hash) framework. (1) In the preprocessing stage, the model picks up similar images from the dataset by labels, and attaches random noise to these images as part of input which is used to get view-relation matrix. The output of the model is $ q $ bits binary code. ( $ b_i \in [-1,1]^q $  while training ) (2) ``Basic $\sim$'' means basic binary code which is generated by basic CNN based hash learning model. (3) View stability evaluation $\mathcal{E}_m(\mathbf{Q})$ calculates view-relation matrix. This part is replaced by memory network during the testing process (step2).}
	\label{graph:newframwork}
\vspace{-0.3cm}
\end{figure*}

In this work, a supervised multi-view hashing method called D-MVE-Hash is proposed for accurate and efficient image retrieval based on multiple visual features, multi-view hash and deep learning.	
Inspiring by multi-modal and cross-modal hashing methods \cite{Jiang_2017_CVPR,hu2018collective}, we excavate multiple view associations and make this potential knowledge dominant to explore the deep integration of the multi-view information.
To this end, we design a view-independent and robustness view-relation matrix which is calculated from the continuously optimized features of each view to better capture the relation among views.
Fig.~\ref{graph:mmrnet} and Fig.~\ref{graph:newframwork} are the illustration of the key idea and model framework. To cope with learning-to-hash, we retain the relaxation and constraint methods similar to \cite{zhu2016deep,cao2017hashnet,chen2018deep}. D-MVE-Hash automatically learns the optimal hash functions through iterative training. After that, we can achieve high-speed image retrieval using the image hash codes which are generated by the trained D-MVE-Hash. In summary, the main contributions of our work are two-fold:

\begin{itemize}
	
	\item We propose a general flexible end-to-end multi-view feature enhancement  framework for image retrieval. This framework can fuse arbitrary view data combination with a unified architecture by fusion methods. The effectiveness of the framework is mainly empowered by a view-independent and robustness view-relation matrix using the fluctuations of different image features under multiple feature views.

	\item Without loss of generality, we comprehensively evaluate the proposed method on three different datasets and implement detailed ablation experiments for endorsing the properties of our D-MVE-Hash. Extensive experiments demonstrate the superiority of the proposed D-MVE-Hash, as compared to several state-of-the-art hash methods in image retrieval task.

\end{itemize}

%
%


\section{Related Work}
\subsection{Approximate Nearest Searching with Hashing}

The image retrieval mentioned in this paper refers to content-based visual information retrieval \cite{banerjee2015using,8662712,8361043}. This process can be simply expressed as: feeding the unlabeled original images to a deep net architecture or other retrieval methods to get images which are similar or belong to the same category of the inputs. It is the typical similarity searching problem. The similarity searching of multimedia data such as collections of images (e.g, views) or 3D point clouds \cite{Qi_2017_CVPR,biyao2020} usually requires the compression processing. Similarity searching (or proximity search) is achieved by means of nearest neighbor finding \cite{arya1998optimal,zhu2019eff,zhu2019PR}. Hashing is an efficient method for nearest neighbor search in large-scale data spaces by embedding high-dimensional feature descriptors into a similarity preserving Hamming space with a low dimension. 

In this paper, we do research on the supervised hash learning algorithm \cite{DBLP:conf/cvpr/ShenSLS15,DBLP:journals/tip/ShenSSHTS15} which uses tagged datasets. Compared with unsupervised hashing algorithm \cite{DBLP:journals/pami/ShenXLYHS18} such as LSH \cite{gionis1999similarity}, KLSH \cite{kulis2009kernelized} and ITQ \cite{gong2012iterative}, supervised methods could obtain more compact binary representaion and generally achieve better retrieval performance \cite{cao2017hashnet,cao2018deep}. LSH \cite{gionis1999similarity} implements a hash retrieval by generating some hash functions (which is called the family $\mathcal{H}$ of functions) with local sensetive properties and applying some random choice methods to transform the input image and construct a hash table. KLSH's \cite{kulis2009kernelized} main technical contribution is to formulate the random projections necessary for LSH \cite{gionis1999similarity} in kernal space. These sensible hashing methods show that the local clustering distance of the constrained binary code plays an important role in the overall optimization. ITQ \cite{gong2012iterative} finds a rotation of zero-centered data so as to minimize the quantization error of mapping this data to the vertices of a zero-centered binary hypercube. KSH \cite{liu2012supervised} maps the data to compact binary codes whose Hamming distance are minimized on similar pairs and simultaneously maximized on dissimilar pairs. All these traditional methods reveal the value of hash learning in retrieval. However, since the deep neural network have shown strong usability in various fields, researchers began to consider introducing the advantages of neural network and convolution operation into hash learning and proposed deep hash learning.

\subsection{Hashing with CNN}

Deep learning based image retrieval methods have shown superior performance compared with the methods using traditional handcraft descriptors. The convolutional neural network \cite{lecun2015deep} is one of the most famous and efficacious deep learning based image processing method for various tasks such as image retrieval and classification. Recently, introducing deep learning into hashing methods yields breakthrough results on image retrieval datasets by blending the power of deep learning \cite{lecun2015deep}. 

CNNH \cite{xia2014supervised} proposed by Xia \textit{et al.} is the first end-to-end framework that used the pairwise similarity matrix and deep convolutional network to learn the hash functions. Xia \textit{et al.} propose a deep convolutional network tailored to the learned hash codes in $H$ and optionally the discrete class labels of the images. The reconstuction error is minimized during training. In \cite{lai2015simultaneous}, Lai \textit{et al.} make the CNNH \cite{xia2014supervised} become an ``one-stage'' supervised hashing method with a Triplet Ranking Loss. Zhu \textit{et al.} \cite{zhu2016deep} proposed the DHN which uses a pairwise cross-entropy loss $L$ for similarity-preserving learning and a pairwise quantization loss $Q$ for controlling hashing quality. HashNet \cite{cao2017hashnet} attacks the ill-posed gradient problem in optimizing deep networks with non-smooth binary activations by continuation method. DMDH \cite{chen2018deep} transforms the original binary optimization into differentiable optimization problem over hash functions through series expansion to deal with the objective discrepancy caused by relaxation.

\subsection{Multi-view Hashing}
\label{sec:relamultivihash}

The poor interpretability of deep learning makes it difficult to further optimize the model in a target manner. We have noticed that some multi-view-based hashing methods \cite{chen2018collaborative,xie2017dynamic,zhang2011composite,liu2015multiview,shen2015multi,zhang2017semi} have emerged in recent years. This type of method processes images by traditional means and extracts various image features for enhancing the deep learning hashing method. The dispersion of hash codes is one of the reasons why such operations are effective. Since the data form among features is very different, it can be seen as viewing images from different views.

The observation of an object should be multi-angled and multi-faceted. Whereas a quantity of multi-view hashing method emphasizes the importance of the multi-view space, ignoring the role of convolution and the relationship among views. Using kernel functions and integrated nonlinear kernel feature is a common way to achieve that, and we can use a weighting vector to constrain different kernels. However, the weighting vector which can gather relationships between views is often pre-set or automatically optimized as a parameter, which is not reasonable enough. In \cite{zhang2011composite}, Zhang \textit{et al.} used graphs of multi-view features to learn the hash codes, and each view is assigned with a weight for combination. Kim \textit{et al.} proposed Multi-view anchor graph hashing \cite{kim2013multi} which concentrates on a low-rank form of the averaged similarity matrix induced by multi-view anchor graph. In \cite{shen2015multi}, Shen \textit{et al.} set $ \mu_m $ measures the weight of the $ m $-th view in the learning process. Multiview Alignment Hashing (MAH) \cite{liu2015multiview} seeks a matrix factorization to effectively fuse the multiple information sources meanwhile discarding the feature redundancy. Xie \textit{et al.} proposed Dynamic Multi-View Hashing (DMVH) \cite{xie2017dynamic} which augments hash codes according to dynamic changes of image, and each view is assigned with a weight. Based on the above, we conducted in-depth research.



\section{Deep Multi-View Enhancement hashing}
\label{sec:dmvhhash}

In this section, the details of our proposed D-MVE-Hash are introduced. We first briefly introduce the definition of multi-view problems and give the detailed description of MV-Hash. Following, we discuss the enhancement process and propose three different data fusion methods. In the last part of the section, We introduce the joint learning and the overall architecture of D-MVE-Hash.


\subsection{Problem Definition and MV-Hash}
\label{sec:dmvhhash1}

Suppose $\mathbf{O}=\{o_i\}^N_{i=0}$ is a set of objects,
and the corresponding features are
$\{\mathbf{X}^{(m)}=[x^{(m)}_1,\cdots,x^{(m)}_N] \in \mathbb{R}^{d_m \times N} \}^M_{m=1}$,
where $d_m$ is the dimension of the m-th view, $M$ is the number of views, and
$N$ is the number of objects.
We also denote an intergrated binary code matrix
$\mathbf{B}=\{b_i\}^N_{i=1} \in \{-1,1\}^{q \times N} $,
where $b_i$ is the binary code associated with $o_i$,
and $q$ is the code length.
We formulate a mapping function $\mathcal{F}(\mathbf{O})=[F_1(\mathbf{X}^{(1)}),\cdots,F_M(\mathbf{X}^{(M)})]$,
where the function $F_m$ can convert a bunch of similar objects into classification scores in different views.
Then, we define the composition of the potential expectation hash function $\varphi :\mathbf{X} \rightarrow \mathbf{B}$ as follow:
\begin{equation}\begin{aligned}
\varphi(\mathbf{X})=[
& \varphi_1(\mathcal{E}(\mathcal{F}(\mathbf{X}_1,\cdots,\mathbf{X}_N)),\mathbf{X}^{(1)}), \cdots ,\\
& \varphi_m(\mathcal{E}(\mathcal{F}(\mathbf{X}_1,\cdots,\mathbf{X}_N)),\mathbf{X}^{(m)})]
\end{aligned},\end{equation}

Before starting stability evaluation, we pre-train each view network in the tagged dataset for classification task.
Using the following loss function:
\begin{equation}\label{equ:clsloss}
\mathcal{L}^p(x, i) = -\log\frac{exp(x[i])}{\sum_j exp(x[j])}.
\end{equation}
The criterion of Eq~\ref{equ:clsloss} expects a class index in the range [0, class-numbers] as the target for each value of a 1D tensor of size mini-batch. The input $x$ contains raw, unnormalized scores for each class. For instance, $x[0]$ is outputted by the classifier to measure the probability that the input image belongs to the first class. $x[1]$ is the prediction score of the second class. $x[i]$ is the prediction score of the ground truth. 
Specific to image data, given $N$ images $\mathbf{I}=\{i_1,...,i_N\}$.
Set $\mathbf{Q}=\mathcal{F}(\mathbf{I})$ in which $\mathcal{F}$ means the testing process. The dimension of $\mathbf{Q}$ is $M\times N\times C$,
where $M$ is the number of views,
$N$ is the number of images,
$C$ is the number of classes.
$Q_{mc}$, which is actually $Q_{mc\cdot}$, omits the third dimension represented by ‘$\cdot$’ and stands for a one-dimensional vector rather than a number.
$\mathcal{E}(\mathcal{F})$ is as follow:
\begin{equation}\label{equ:core}\begin{aligned}
\mathcal{E}_m(\mathbf{Q})= 
& +\sum_{m=1}^M \max_c \sqrt{\mathcal{V}(\mathbf{Q}_{mc})}\\
& -\frac{1}{N} \sum_{c=1}^C \sqrt{\mathcal{V}(\mathbf{Q}_{mc})} 
\end{aligned},\end{equation}
and the $\mathcal{V}(\mathbf{Q}) = \frac{1}{N} \sum_{n=1}^N(\mathbf{Q}-\mu)$,
where the $\mu$ is arithmetic mean.
$\mathcal{E}$ is expressed as $[\mathcal{E}_1,\cdots,\mathcal{E}_M]$.
Then we do a simple normalization of $\mathcal{E}$:
\begin{equation}
\log\{\mathcal{E}_m(\mathbf{Q})\}=\frac{\log (\mathcal{E}_m(\mathbf{Q})+\vert\min(\mathcal{E}(\mathbf{Q}))\vert +1)}{\log (\max (\vert\mathcal{E}(\mathbf{Q})\vert)+\vert\min(\mathcal{E}(\mathbf{Q}))\vert+1)}.
\end{equation}

Then we consider training multi-view binary code generation network with view-relation information.
At the beginning, we ponder the case of a pair of images $i_1,i_2$ and corresponding binary network outputs $b_1,b_2 \in \mathbf{B}$, which can relax from $ \{-1,+1\}^q $ to $[-1,+1]^q$ \cite{weiss2009spectral}. We define $y=1$ if they are similar, and $y=-1$ otherwise.  The following formula is the loss function of the m-th view:
\begin{equation} 
\label{equ:7}
\begin{aligned}
\mathcal{L}_m(b_1, b_2,y)= -
& (y-1)\ \max (a-\Vert b_1-b_2\Vert^2_2,0)\\+
& (y+1)\Vert b_1-b_2\Vert^2_2\\+
& \alpha (\Vert\vert b_1 \vert -1 \Vert_1 + \Vert\vert b_2 \vert -1 \Vert_1)
\end{aligned},
\end{equation}
where the $\Vert\cdot\Vert_1$ is the L1-norm, $\vert\cdot\vert$ is the absolute value operation, $\alpha>0$ is a margin threshold parameter, and the third term in Equation~\ref{equ:7} is a regularizer term which is used to avoid gradient vanishing. 

More generally, we have the image input $\mathbf{I}=\{i_1,...,i_N\}$ and output $\mathbf{B}^{(m)}=\{b_1^{(m)},...,b_N^{(m)}\}$ in the multi-view space. In order to get the equation representation in matrix form, we substitute $\mathbf{B}$ into $\mathcal{L}_m$ formula, then complement the regular term and similarity matrix to get following global objective function:
\begin{equation}
\label{equ:9}
\begin{array}{c}\vspace{0.1cm}
\mathcal{L}(\mathbf{I},\mathbf{Y})=\\
\mathbf{Y}_{ij} \cdot \log\{\mathcal{E}_m(\mathcal{F}(\mathbf{I})\} \cdot
(\rho(\mathbf{I})+\alpha\Vert\varphi(\mathbf{I})'-\mathbf{1}_{K_{N*q}}\Vert)
\end{array}.\end{equation}
The matrix merged by $\varphi(\mathbf{I})$ is denoted as $\rho(\mathbf{I})$ refer to the form of the second term of multiplication in Equation~\ref{equ:8}. The $\delta(\cdot)$ in Equation~\ref{equ:8} is the $maximum$ operation. The $\oplus$ is inner product.
\begin{equation} 
\label{equ:8}
\begin{aligned}
\left[\begin{array}{c}
y(b_i,b_j)=+1\\
y(b_i,b_j)=-1
\end{array}\right]
\cdot
\left[\begin{array}{c}
\delta((\mathbf{B}\oplus\mathbf{B})\cdot(\mathbf{B}\oplus\mathbf{B})')\\
(\mathbf{B}\oplus\mathbf{B})\cdot(\mathbf{B}\oplus\mathbf{B})'
\end{array}\right]
\end{aligned}.\end{equation}

In order to show the effect of view stability evaluation in Equation~\ref{equ:9} intuitively, we rewrite the overall loss function as Equation~\ref{equ:10}. That is, we want to highlight the position of view-relation matrix $\mathbf{E}$ which is the output of $\mathcal{E}(\mathcal{F}(\mathbf{I}))$ in the overall loss function. 
\begin{equation}\label{equ:10} \begin{aligned}
\mathcal{L}(\mathbf{E},\mathbf{B})=
\sum_{n=1}^N \sum_{m=1}^M \mathbf{E}_{nm} \sum_{i=1}^N \mathcal{L}_m(\mathbf{B}_n^{(m)},\mathbf{B}_i^{(m)},y)
\end{aligned}.\end{equation}
With this objective function, the network is trained using back-propagation algorithm with mini-batch gradient descent method. Meanwhile, since the view-relation matrix $\mathbf{E}$ is directly multiplied by $\mathcal{L}_m$ in Equation~\ref{equ:10}, the view relationship information can affect the direction of gradient descent optimization.

\subsection{Enhancement and Fusion}

\begin{figure}[t]
	\centering
	\includegraphics[scale=0.47]{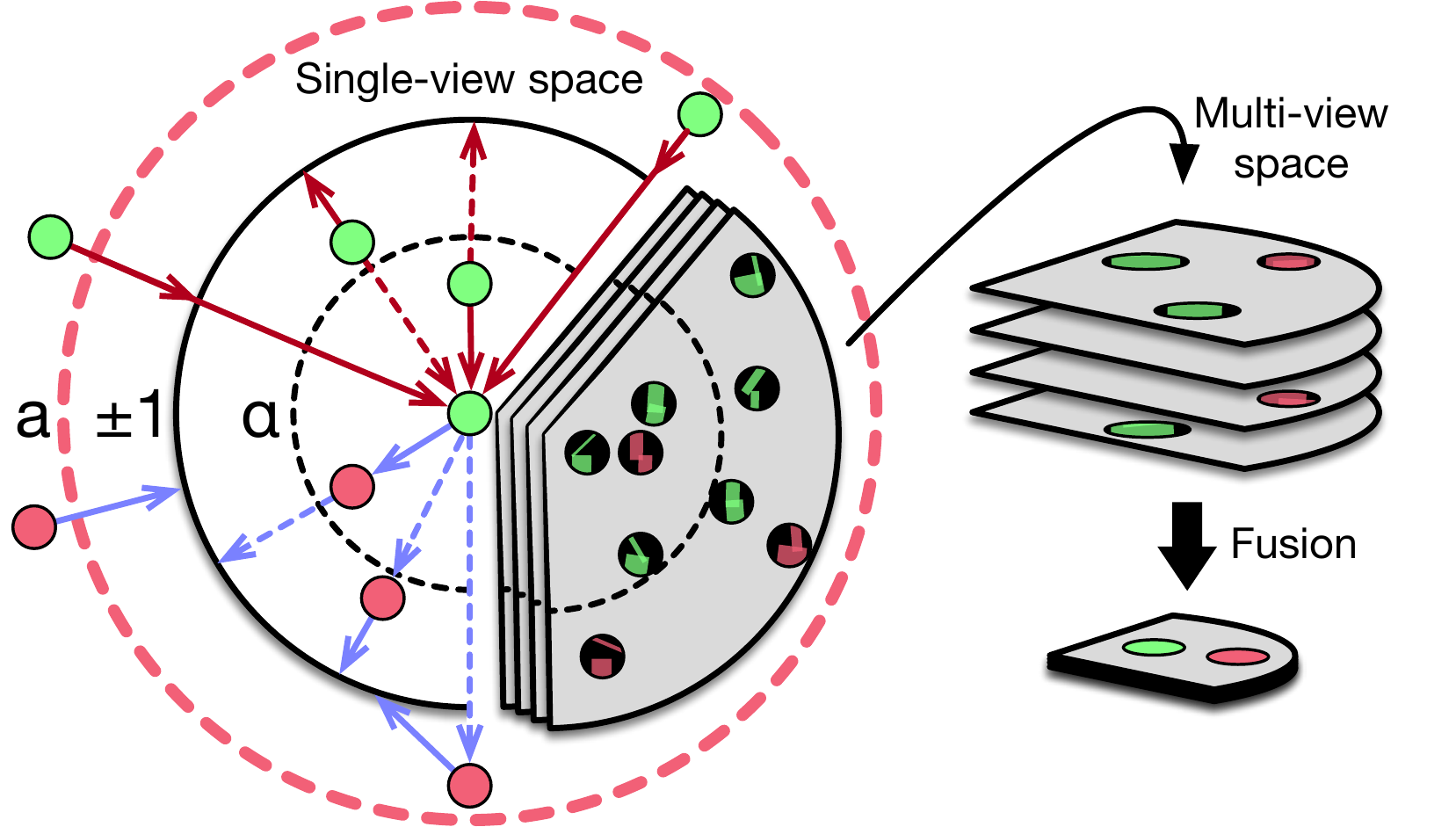}
	\caption{Illustration of the data enhancement process.The left half of the figure shows the double-sample hash mapping constraint rules (e.g., Equation~\ref{equ:7}) which are used to embed features into the identical Hamming space.}
	\label{graph:loss}
\vspace{-0.3cm}
\end{figure}

The next stage is integrating multi-view and view-relation information into the traditional global feature in the Hamming space. Fig.~\ref{graph:loss} is the illustration of the data enhancement process which is actually divided into two parts. Firstly, we use some hash mapping constaint rules which are occur simultaneously in single-view and multi-view spaces to embed features into the identical Hamming space. Secondly, we propose replication fusion, view-code fusion and probability view pooling which are all carried out in the Hamming space to accomplish the multi-view data, view-relation matrix and traditional global feature integration.

In Fig.~\ref{graph:loss}, the solid and dotted line respectively indicate the strong and weak constraint. For example, as for the green points: 1) The distance constraint (i.e., the clustering) is stronger than $\pm 1$ constraint (i.e., the hash dispersion) when points are within the $\alpha$ circle. Therefore, we connect the green points by solid lines, and connect the green dots with $\pm 1$ circle by dotted lines; 2) The distance constraint is weaker than $\pm 1$ constraint when points are between the $\alpha$ circle and $\pm 1$ circle. Therefore, we connect the green points by dotted lines, and connect the green dots with $\pm 1$ circle by solid lines; 3) The distance constraint and $\pm 1$ constraint are both the strong constraints when points are outside the $\pm 1$ circle. Similar operations also occur on the red points. Compared to the green points, the clustering direction of the red points is opposite.

\label{sec:dmvhhash2}


\subsubsection{Replication Fusion}
\label{sec:f1}

Replication fusion is a relatively simple solution which relies on parameters. We sort $\mathbf{E}$ to find important views, and strengthen the importance of a view by repeating the binary code of the corresponding view in multi-view binary code. 
Specifically, the basic binary code is denoted as $\mathbf{B}$. The intermediate code (multi-view binary code) is denoted as $\mathcal{H}$. We set fusion vector $v$ to guide the repetition of multi-view binary code under various views. Equation~\ref{equ:11} represents the encoding process.
\begin{equation}
\label{equ:11}
\begin{aligned}
\mathbf{H}= [\mathbf{B}\ ,\ \mathop{\phi}\limits_{j=1}^M(\mathop{\phi}\limits_{i=1}^M((\mathcal{H})_i,v_i\cdot \mathop{\mathcal{S}}\limits_{d=0}(\mathbf{E}))_j,1)]
\end{aligned},\end{equation}
where $\mathbf{H}$ represents the input binary code of the fusion layer. $\phi(\cdot)$ from $1$ to $M$ is the self-join operation of the vector. The second parameter in $\phi(\cdot)$ represents the number of self-copying. $\mathcal{S}$ is a sort function in $d$ dimension. The advantage of this fusion method is that it can convert $\mathbf{E}$ into a discrete control vector, therefore $\mathbf{E}$ only determines the order between views. The strength of enhancement or weakening is manually controlled by fusion vector. 

\subsubsection{View-code Fusion}
\label{sec:f2}

View-code fusion considers the most primitive view-relation matrix and the least artificial constraints. Specifically, we want to eliminate fusion vector which is used to ensure that the dimensions of input data are unified in Fusion-R because of the dynamic view-relation matrix. At first, the entire binary string $\mathbf{H}$ is encoded into head-code ($\mathbf{H}_h$), mid-code ($\mathbf{H}_m$), and end-code ($\mathbf{H}_e$). $\mathbf{H}_h$ is the same as Fusion-R. $\mathbf{H}_m$ directly uses the product of binary code length and the coefficient of corresponding view as the current code segment repetition time. This operation produces a series of vacant bytes (i.e., $\mathbf{H}_e$) which are not equal in length. Second, we assign a specific and different coden called view-code which is a random number belonging to $[-1, 1]$ in each view. Compared with $\mathbf{H}_m$, $\mathbf{H}_e$ uses view-code instead of multi-view binary code. So that it can be completely filled regardless of the dynamic view-relation matrix and code length. The advantage of view-code fusion is that it fully utilizes the information contained in view-relation matrix. We find that view-code fusion is limited by view stability evaluation in our experiments, which means it can exceed replication fusion when the number of views increases.

\subsubsection{Probability View Pooling}
\label{sec:f3}

We propose probability view pooling with view-relation matrix as a multi-view fusion method. Traditional pooling operation selects maximum or mean as the result of each pooling unit. The view pooling is a dimensionality reduction method which use element-wise maximum operation across the views to unify the data of multiple views into one view \cite{feng2018gvcnn}. Since pooling operation inevitably cause information loss, we need to expand the length of multi-view binary code to retain multi-view information as much as possible before probability view pooling. Then the view probability distribution is generated according to $\mathbf{E}$. In each pooling filter, a random number sampled from the view probability distribution activates the selected view. The code segment of this view is used for traditional pooling operation. It ensures that sub-binaries of high-priority views are more likely to appear during the fusion process.


\subsection{Joint Learning}
\label{sec:optim}

In this section, we introduce a multi-loss synergistic gradient descent optimization for the proposed model. D-MVE-Hash is optimized based on $\mathcal{L}^p$ and $\mathcal{L}^c$ at first. The former is apply for view stability evaluation, and the latter is used to extract the basic binary code. At this stage, our loss function is $\mathcal{L}^p + \mathcal{L}^c$. Then we use $\mathcal{L}$ to train the backbone of D-MVE-Hash (including fusion part) and use $\mathcal{L}^w$ which is explained in the last paragraph of this section to train the memory network. At this stage, our loss function is $\mathcal{L} + \mathcal{L}^w$. As a consequence, the formula for segment optimization is as follows:
\begin{equation} \begin{aligned}
\min_\Theta\mathcal{L}=\left\{
\begin{array}{l}\vspace{0.1cm}
\mathcal{L}_1=\mathcal{L}^p + \mathcal{L}^c\\
\mathcal{L}_2=\mathcal{L}\ \ +\mathcal{L}^w
\end{array}
\right.
\end{aligned}.\end{equation}
With the purpose of avoid losing gradient, all segmentation maps are not allowed, therefore the output is controlled within $[-1,1]$ (by using the third item in original $\mathcal{L}$). Finally, the $sgn(\cdot)$ function is used to map the output to the Hamming space during testing, thus we can measure the Hamming distance of the binary code.

We also introduce the memory network into the proposed D-MVE-Hash to avoid excessive computing resources of the view stability evaluation method. It is noted that the memory network is a simplification of the view stability evaluation method. The memory network only focuses on the original image input and the view-relation matrix. It can learn such transformation through multiple iterations during training. However, since we use pre-converted hash codes for retrieval, the complexity of the model structure and the time complexity of the retrieval are separate. This means D-MVE-Hash still maintains the inherent advantages of hashing which is a very important and definitely the fastest retrieval method since such Hamming ranking only requires O(1) time for each query.

More specifically, the memory network learns the view-relation matrix $\mathbf{E}$ in setp1, and then in step2, we can get view-relation matrix $\mathbf{E}$ by this module without using stability evaluation method. The structure of memory network is a multi-layer convolutional neural network (e.g. VGG, ResNet, DenseNet, etc.), but its output layer is relative to view-relation matrix $\mathbf{E}$. And the loss function during training is $\mathcal{L}^w = \{l_1,\dots,l_N\}^\top$, $l_n = \left( I_n - \mathbf{E}_n \right)^2$. Fig.~\ref{graph:mmrnet} shows the different states and association of D-MVE-Hash between two steps. In general, we design such a structure mainly for engineering considerations rather than performance improvement. In our actual model training, whether or not the memory network is used does not have much influence on retrieval performance.

\begin{figure}[t]
	\centering
	\subfloat[Same Hamming radius]{
		\includegraphics[scale=0.17]{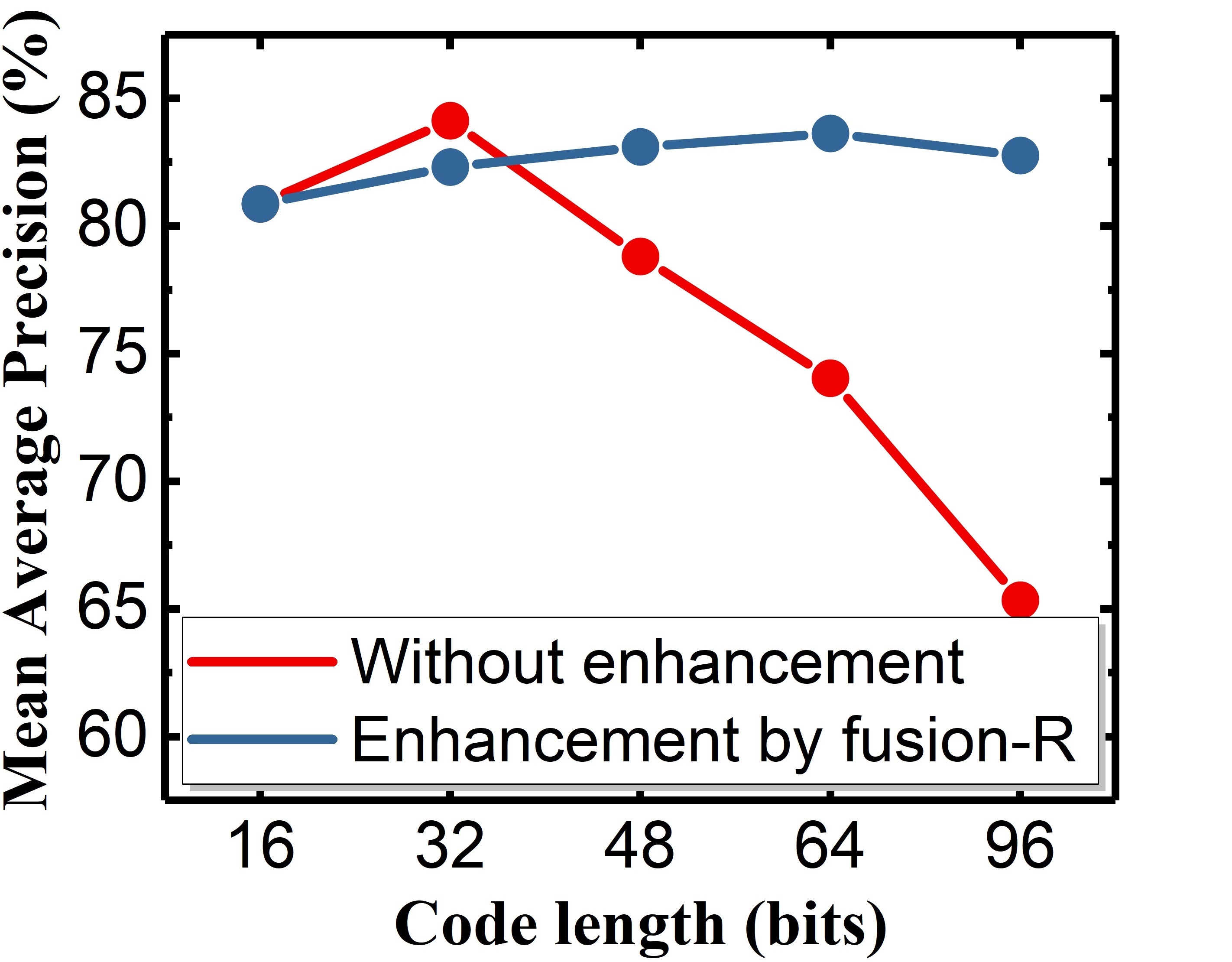}\label{graph:lencifar}
	}
	\subfloat[Different Hamming radius]{
		\includegraphics[scale=0.17]{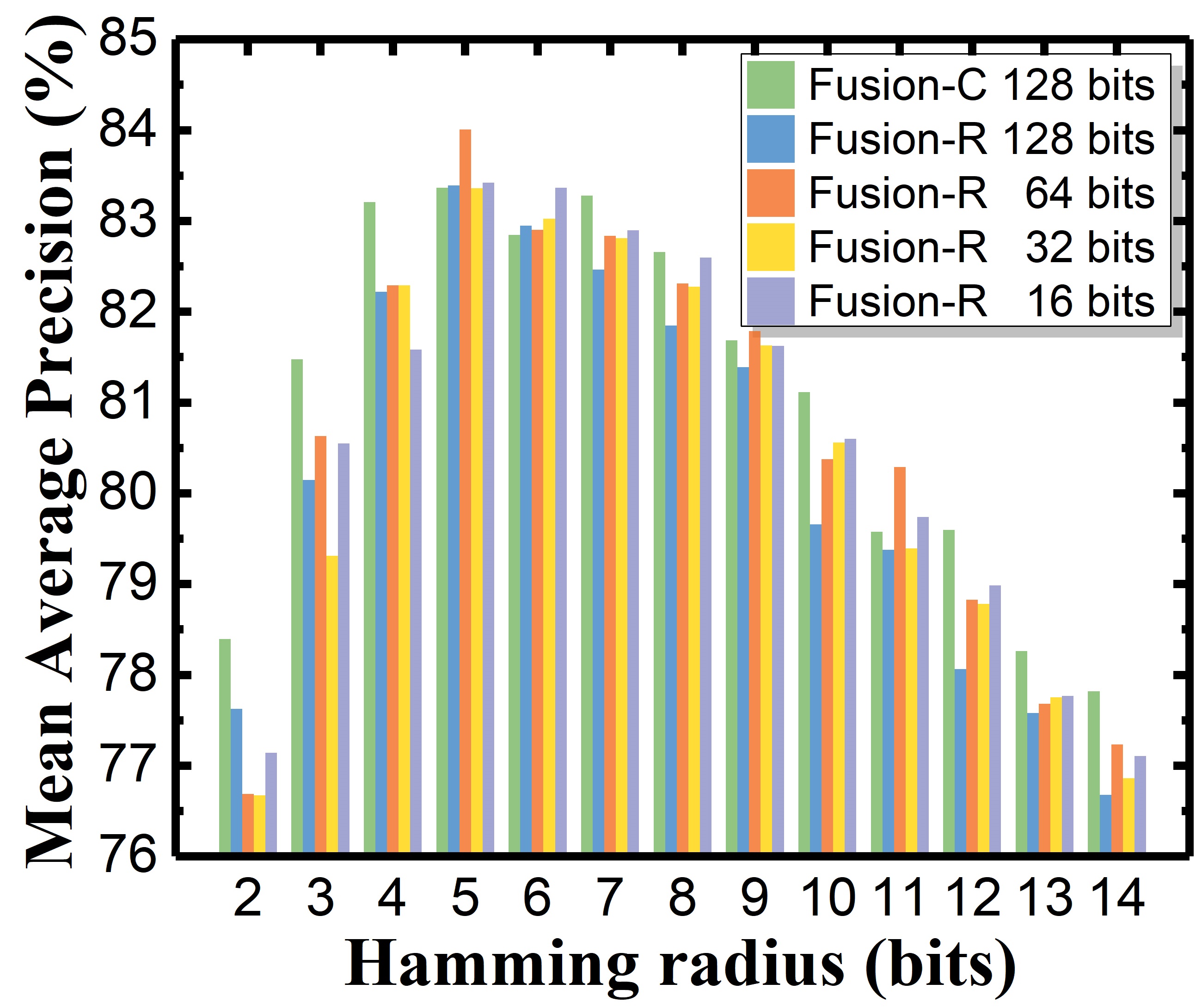}\label{graph:radiu}
	}
	\caption{(a) Retrieval results of single-view hashing method and the proposed multi-view hashing method with the same Hamming radius. (b) The mean average precision (mAP) for different code lengths and Hamming radius on CIFAR-10 dataset.}
	\label{graph:len}
	\vspace{-0.6cm}
\end{figure}

\begin{figure}[t]
	\centering
	\subfloat[Replication Fusion]{
		\includegraphics[scale=0.16]{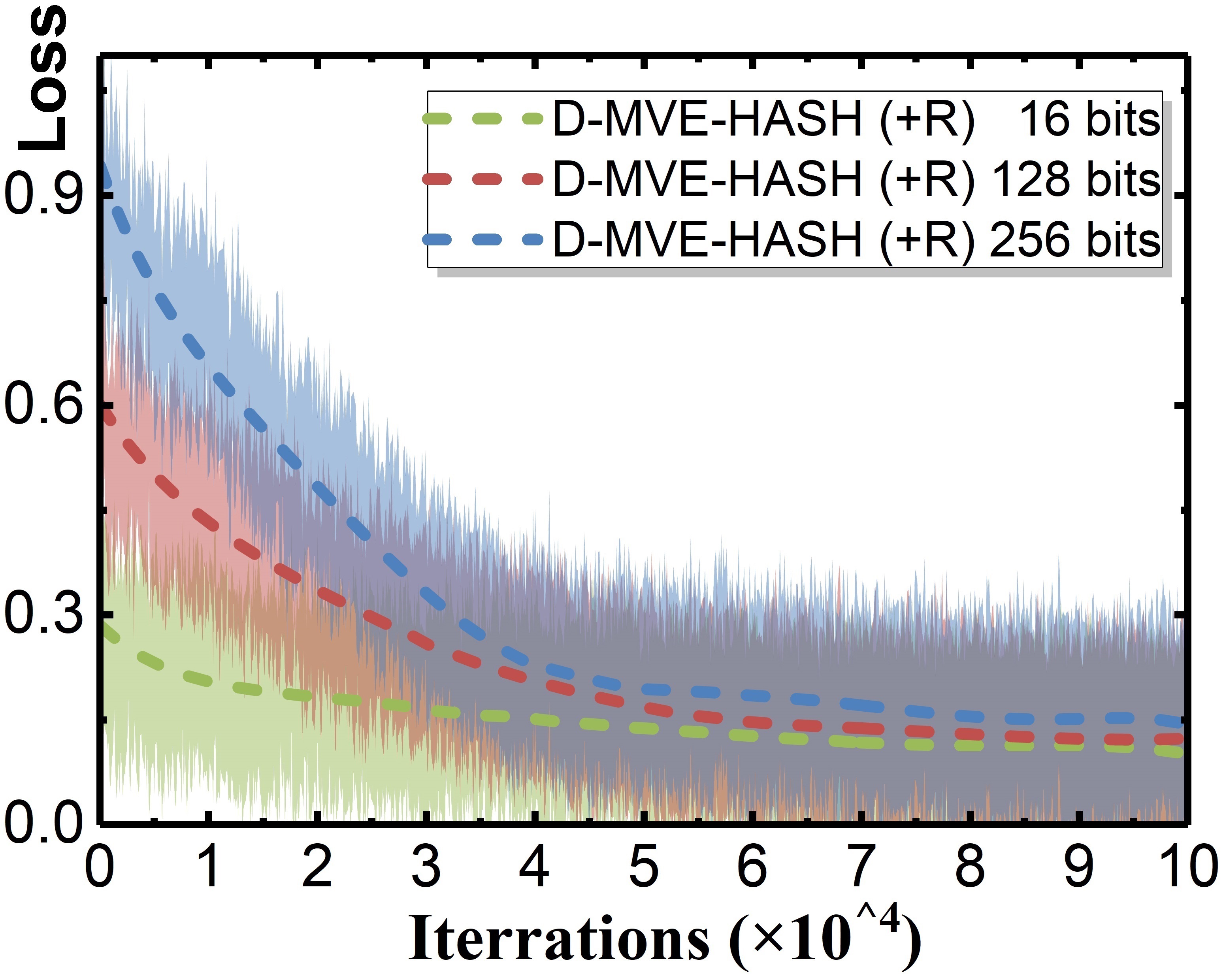}\label{graph:lossfusionr}
	}
	\subfloat[View-Code Fusion]{
		\includegraphics[scale=0.16]{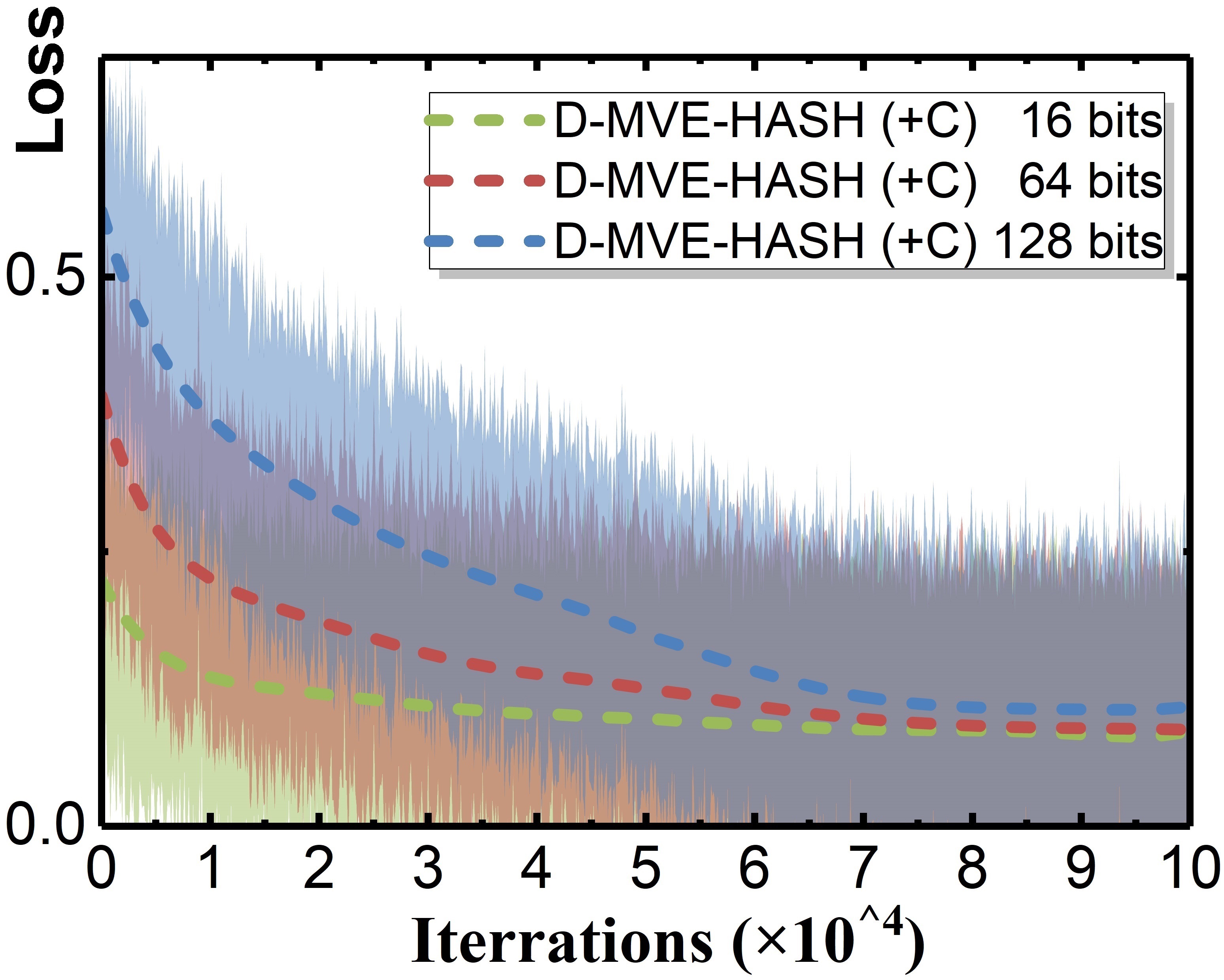}\label{graph:lossfusionc}
	}
	\caption{(a) The D-MVE-Hash (+R) training loss with 16 bits, 128 bits and 256 bits hash codes. (b) The D-MVE-Hash (+C) training loss with 16 bits, 128 bits and 256 bits hash codes.}
	\label{graph:lossfusion}

\end{figure}

\begin{figure}[t]
	\centering
	\subfloat[On CIFAR-10 dataset]{
		\includegraphics[scale=0.17]{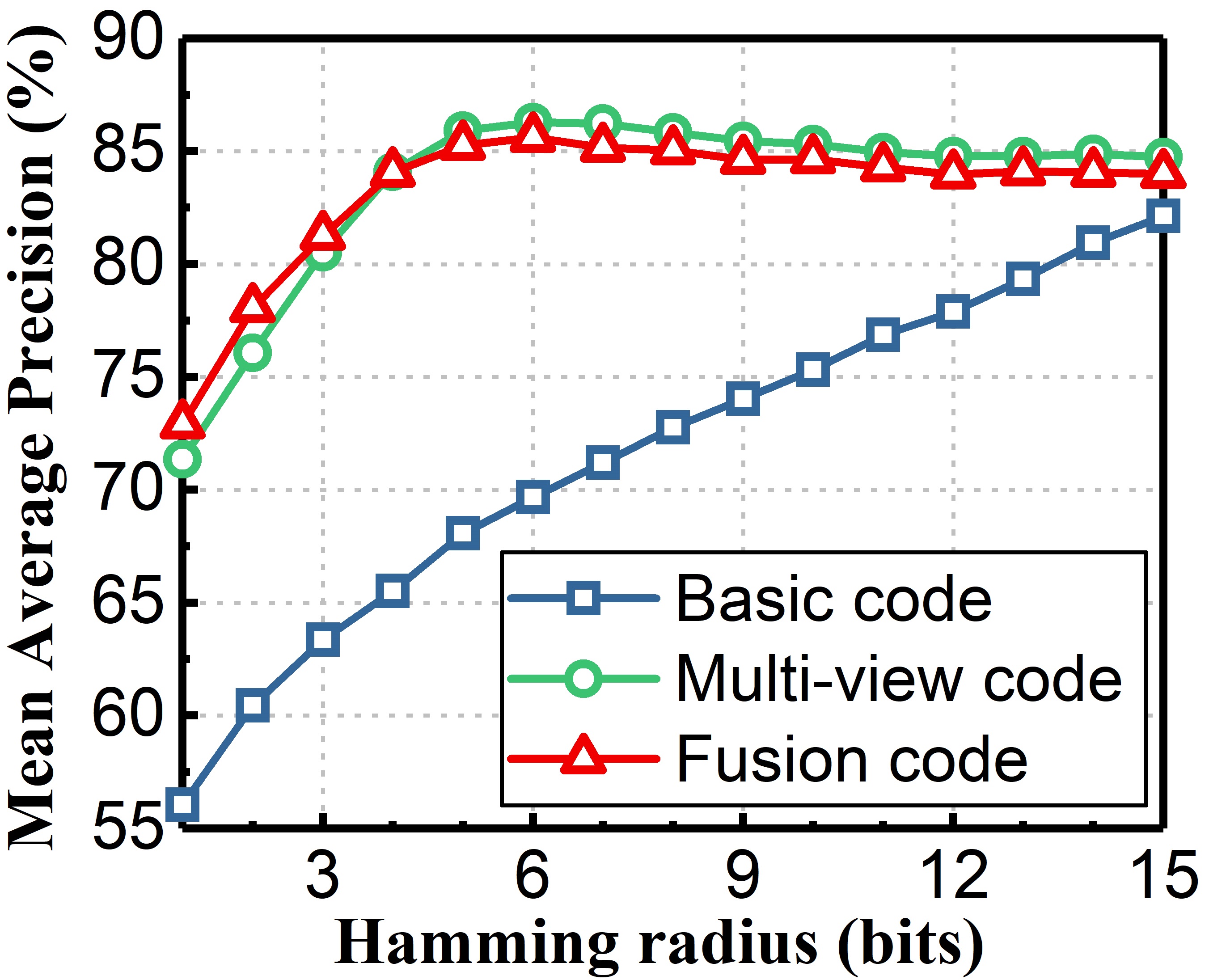}\label{cifarradiu}
	}
	\subfloat[On NUS-WIDE dataset]{
		\includegraphics[scale=0.17]{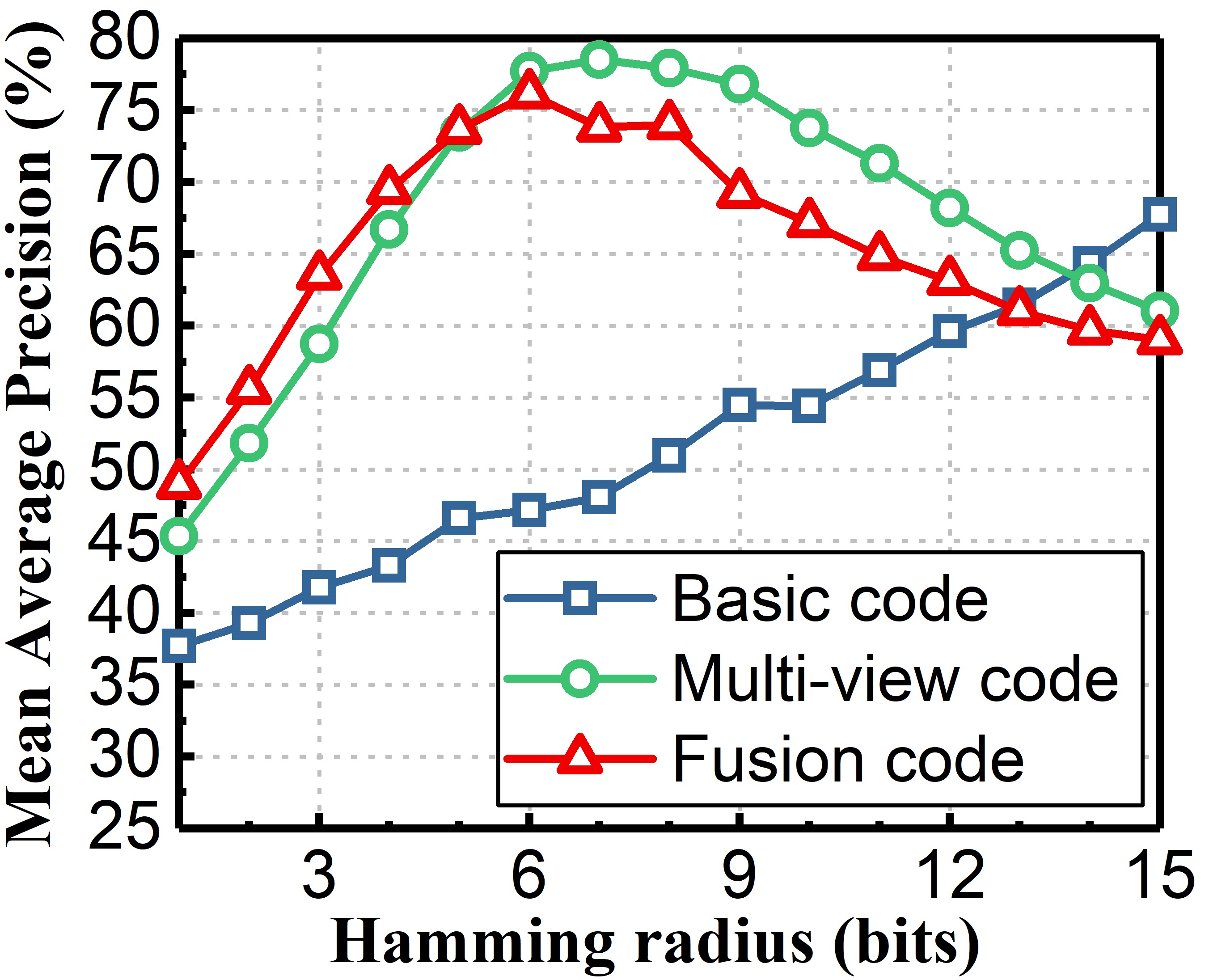}\label{nusradiu}
	}
	
	\caption{The mean average precision (mAP) for the single-view hashing method (the outputs is the basic binary codes), MV-Hash (the outputs is the multi-view binary codes) and  D-MVE-Hash (the outputs is the fusion binary codes) using 96 bits hash code.}
	\label{graph:doubleradiu}
	\vspace{-0.6cm}
\end{figure}

\begin{figure}[t]
	\centering
	\subfloat[On CIFAR-10 dataset]{
		\includegraphics[scale=0.17]{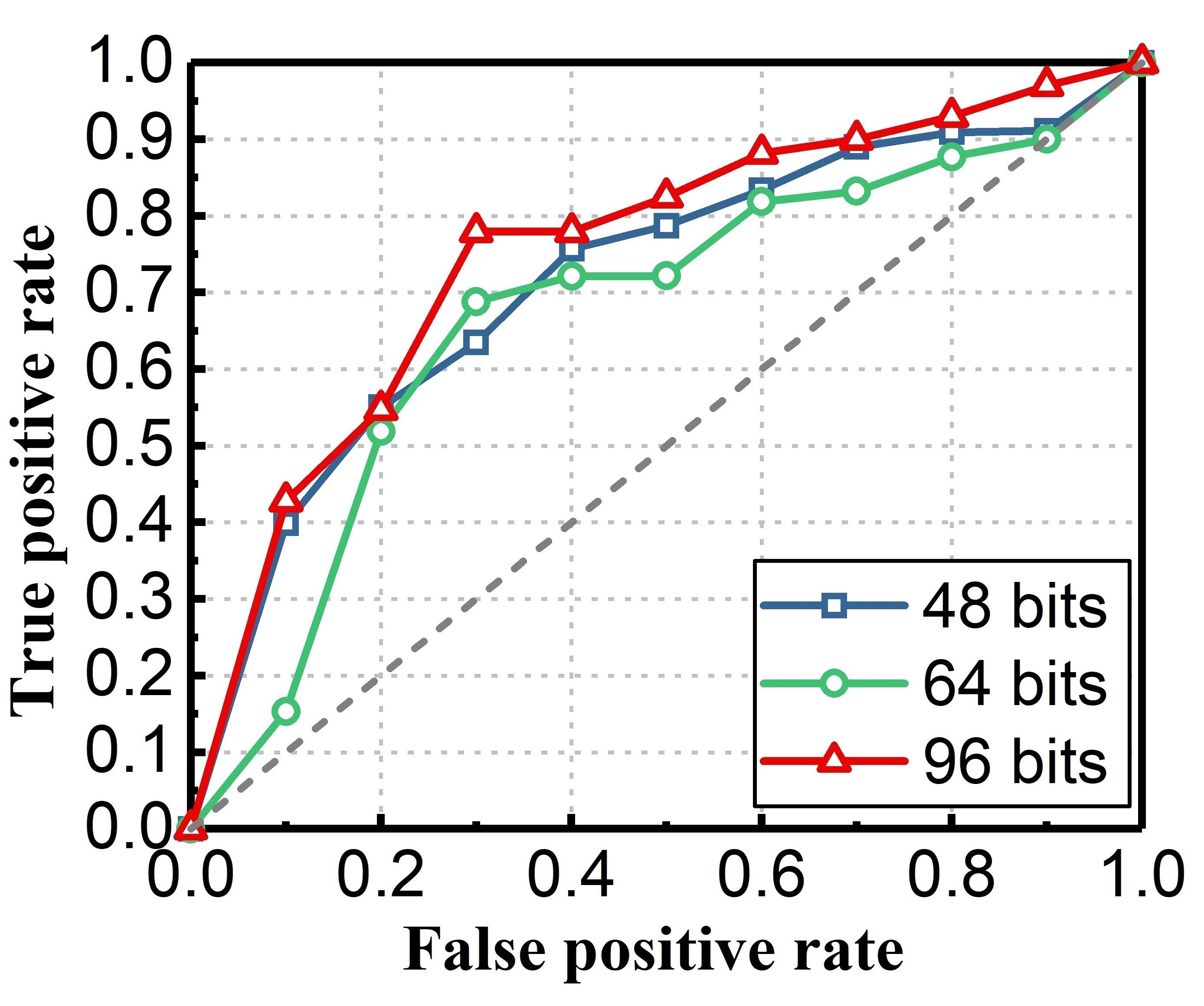}\label{cifarroc}
	}
	\subfloat[On NUS-WIDE dataset]{
		\includegraphics[scale=0.17]{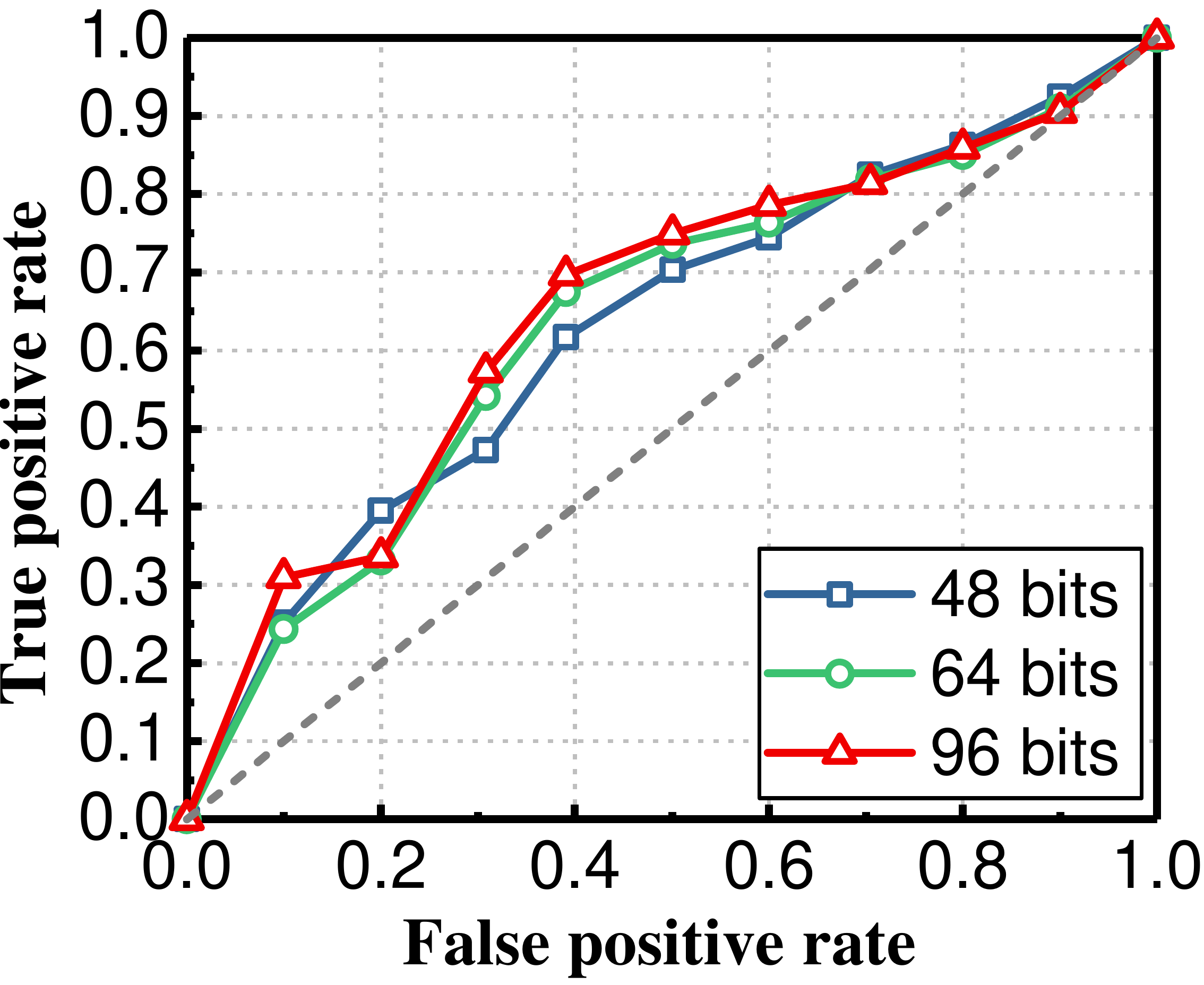}\label{nusroc}
	}
	\caption{The receiver operating characteristic curves (ROC) for the fusion binary codes using 48 bits, 64 bits and 96 bits hash codes.}
	\label{graph:roc}
		\vspace{0.4cm}
\end{figure}

\section{Experiments}
\label{sec:exper}

In this section, we provide experiments on several public datasets, and compare with the state-of-the-art hashing methods. Multi-view hashing methods is also within our scope of comparison. Following \cite{liu2015multiview}, we obtain the 2D images multi-view information through RGB color space color histogram, HSV color space color histogram, texture \cite{ojala2002multiresolution}, and hog \cite{dalal2005histograms}. The implementation of D-MVE-Hash is based on PyTorch 1.1.0 and scikit-image 0.13.1 framework. Each fully connection layer uses Dropout Batch Normalization \cite{srivastava2014dropout} to avoid overfitting. The activation function is ReLu and the size of hidden layer is $4096 \times 4096$. We use mini-batch stochastic gradient decent (SGD) with 0.9 momentum. The learning rate is 0.001. The parameter in Eqn.(\ref{equ:7}): $a=2$ and $\alpha=0.01$, and parameter $v=(1,4,8,16)$. Following the standard retrieval evaluation method \cite{cao2018deep} in the Hamming space, which consists of two consecutive steps: (1) Pruning, to return data points within Hamming radius 2 for each query using hash table lookups; (2) Scanning, to re-rank the returned data points in ascending order if their distances to each query using continuous codes.

Three image datasets are used to evaluate our approach: CIFAR-10 \cite{krizhevsky2009learning} ,NUS-WIDE \cite{nus-wide-civr09} and MS-COCO \cite{lin2014microsoft}.
CIFAR-10 consists of 60,000 $32 \times 32$ color images in 10 classes, with 6,000 images per class. The dataset is split to training set of 50,000 images and testing set of 10,000 images.
NUS-WIDE includes a set of images crawled from Flickr, together with their associated tags, as well as the ground-truth for 81 concepts for these images. We randomly sample 4,000 images as training set, 1,000 images as test query. MS-COCO is composed of 82,783 training images and 40,504 validation images, as well as 80 semantic concepts.


\renewcommand\arraystretch{0.2}
\begin{table}[b]
	\vspace{-0.3cm}
	\centering
	\caption{Ablation experiments (64 bits, +R, CIAFR-10). Time(h) is the time for training process (Batch-size 2). The GPU is NVIDIA GeForce GTX 1080Ti 1481-1582MHz 11GB (CUDA 9.0.176 cuDNN 7.1.2)}
	\label{tab:ablation}	
	\begin{tabular}{c|c|c|c|c|c|c}
		\toprule [0.3ex]
		$\mathbf{E}$&view1 & view2 & view3 & view4 & Time(h) & mAP\\ 
		\midrule
		& &  &  &  & 0.954 & 0.7588\\		
		& \checkmark & &  &  & 1.399 & 0.7513\\
		& & \checkmark &  &  & 1.413 & 0.7699\\
		& &  & \checkmark &  & 1.390 & 0.7551\\
		& &  &  & \checkmark & 1.374 & 0.7854\\
		\hline
		\checkmark & \checkmark & \checkmark &  & & 1.664 & 0.8275\\
		\checkmark & \checkmark &  & \checkmark & & 1.659 & 0.8214\\
		\checkmark & \checkmark &  &  & \checkmark& 1.629 & 0.8244\\
		\checkmark & & \checkmark & \checkmark &  & 1.639 & 0.8229\\
		\checkmark & & \checkmark &  & \checkmark & 1.613 & 0.8268\\
		\checkmark & &  & \checkmark & \checkmark & 1.659 & 0.8315\\
		\checkmark & \checkmark & \checkmark & \checkmark &  & 1.659 & 0.8258\\
		\checkmark & \checkmark & \checkmark &  & \checkmark & 1.657 & 0.8237\\
		\checkmark & \checkmark &  & \checkmark & \checkmark & 1.666 & 0.8319\\
		\checkmark & & \checkmark & \checkmark & \checkmark  & 1.672 & 0.8352\\
		\midrule
		\checkmark & \checkmark & \checkmark & \checkmark & \checkmark & 1.711 & 0.8401\\
		\bottomrule [0.3ex]
	\end{tabular}
\end{table}

\renewcommand\arraystretch{1}
\begin{table*}[]
	\centering
	\caption{The mean average precision (mAP) of re-ranking for different bits on the CIFAR-10 (left), NUS-WIDE (middle) and MS-COCO (right) dataset. D-MVE-Hash (+R) means the proposed model D-MVE-Hash with replication fusion enhancement method. D-MVE-Hash (+C) means D-MVE-Hash with view-code fusion enhancement method. D-MVE-Hash (+P) means D-MVE-Hash with probability view pooling enhancement method.}
	\begin{tabular}{c|cccc|cccc|cccc}
		\toprule [0.3ex]
		\\[-2.8ex]
		\multirow{2}{*}{Methods} & \multicolumn{4}{c|}{CIFAR-10} & \multicolumn{4}{c|}{NUS-WIDE} & \multicolumn{4}{c}{MS-COCO} \\[-2.7ex]\\\cline{2-13}\\[-2.2ex]
		& 16 bits& 32 bits  &48 bits & 64 bits & 16 bits& 32 bits  &48 bits & 64 bits & 16 bits& 32 bits  &48 bits & 64 bits\\	
		[-0.9ex]\midrule\\[-2.9ex]
		KSH \cite{liu2012supervised}
		&0.4368 &0.4585 &0.4012 &0.3819 &0.5185 &0.5659 &0.4102 &0.0608 &0.5010 &0.5266 &0.5981 &0.5004\\
		MvDH \cite{shen2018multiview}
		&0.3138 &0.3341 &0.3689 &0.3755 &0.4594 &0.4619 &0.4861 &0.4893 &0.5638 &0.5703 &0.5912 &0.5952\\
		CMH \cite{chen2018collaborative}
		&0.4099 &0.4308 &0.4411 &0.4841 &0.5650 &0.5653 &0.5813 &0.5910 &0.5126 &0.5340 &0.5455 &0.6034\\			
		CNNH \cite{xia2014supervised}
		&0.5512 &0.5468 &0.5454 &0.5364 &0.5843 &0.5989 &0.5734 &0.5729 &0.7001 &0.6649 &0.6719 &0.6834\\
		HashNet \cite{cao2017hashnet}
		&0.7476 &0.7776 &0.6399 &0.6259 &0.6944 &0.7147 &0.6736 &0.6190 &0.7310 &0.7769 &0.7896 &0.7942\\
		DCH \cite{cao2018deep}
		&0.7901 &0.7979 &0.8071 &0.7936 &0.7401 &0.7720 &0.7685 &0.7124 &0.8022 &0.8432 &0.8679 &0.8277\\
		AGAH \cite{gu2019adversary}   	
		&0.8095 &0.8134 &0.8195 &0.8127 &0.6718 &0.6830 &0.7010 &0.7096 &0.8320 &0.8352 &0.8456 &0.8467\\
		[-0.7ex]\midrule\\[-3.1ex]
		\textbf{MV-Hash} &0.5061 &0.5035 &0.5339 &0.5370 &0.6727 &0.6836 &0.7141 &0.7150 &0.5902 &0.5941 &0.6268 &0.6339\\	
		\textbf{D-MVE-Hash (+P)}   &0.7234 &0.7535 &0.7982 &0.7712 &\underline{0.7157} &0.7271 &0.7281 &0.7317 &0.7358 &0.7782 &0.7924 &0.8062\\
		\textbf{D-MVE-Hash (+C)}   &\textbf{0.8422} &\underline{0.8438} &\underline{0.8460} &\underline{0.8346} &0.6906 &\underline{0.7751} &\underline{0.7860} &\underline{0.7938} &\textbf{0.8552} &\textbf{0.9023} &\underline{0.8726} &\underline{0.8623} \\			
		\textbf{D-MVE-Hash (+R)}   &\underline{0.8345} &\textbf{0.8336} &\textbf{0.8501} &\textbf{0.8401} &\textbf{0.7883} &\textbf{0.8002} &\textbf{0.8057} &\textbf{0.8073} &\underline{0.8323} &\underline{0.8483} &\textbf{0.8735} &\textbf{0.8892}\\[-0.2ex]				
		\bottomrule [0.3ex]
	\end{tabular}
	\label{tabl:cnnbasemapcomp}
	\vspace{-0.3cm}
\end{table*}

\subsection{Ablation Experiment for View-relation Matrix $\mathbf{E}$}
\label{sec:exper3}

In this section, we evaluate the D-MVE-Hash in the absence of views and view-relation matrix. As can be seen from the results, since we adopt dominant view relations rather than implication relations, preserving view relationships by generating partial views with the broken $\mathbf{E}$ is feasible. Our D-MVE-Hash does not degenerate into normal image hashing when multi-view information is partially removed. Another special property is robustness. For the purpose of that, we obtain the volatility matrix which regulates the view relevance based on fluctuation strength to produce a quality view-relation matrix. Aditionaly, we use the mathematics (Equation~\ref{equ:core}) to reduce the interference of irrelevant views on the matrix $\mathbf{E}$. In ablation experiments, we degenerate D-MVE-Hash into a non-multi-view method as the baseline for subsequent comparisons. Then we evaluate our model under partial views with the broken $\mathbf{E}$. At last, we use the intact model which achieves gains of 8.13\% when the complete matrix $\mathbf{E}$ is used, compared with the baseline. For the time complexity, our model takes about 1.711(h) to train our D-MVE-Hash with the complete multi-view information and view-relation calculation, which is 0.757(h) slower than the baseline.

\subsection{Comparisons and Retrieval Performance}
\label{sec:exper2}

\begin{figure}[t]
	\centering
	\includegraphics[scale=0.85]{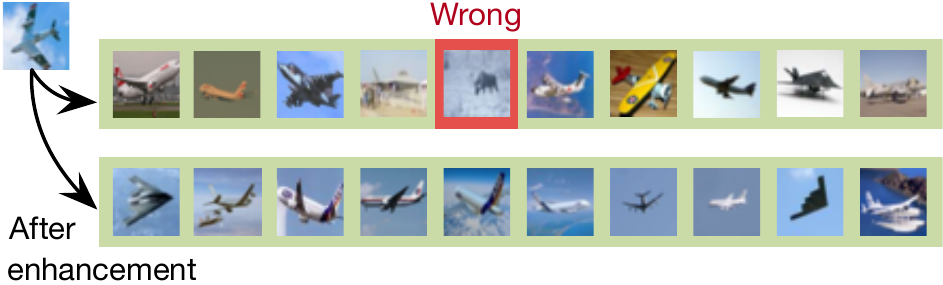}
	\caption{The top 10 retrieved images before and after enhancement.}
	\label{graph:visual}
	\vspace{-0.3cm}
\end{figure}

We compare the retrieval performance of D-MVE-Hash and MV-Hash with several classical single/multi-view hashing method: KSH \cite{liu2012supervised}, MvDH \cite{shen2018multiview},  CMH\cite{chen2018collaborative}, and recent deep methods: CNNH \cite{xia2014supervised}, HashNet \cite{cao2017hashnet}, DCH \cite{cao2018deep} and AGAH \cite{gu2019adversary}. All methods in the Tab.~\ref{tabl:cnnbasemapcomp} are supervised. As shown in these results, we have following observations:

\begin{itemize}
	\item Compared with classical hashing methods, for example, MV-Hash obtains gains of \textbf{9.62}\%, \textbf{7.27}\%, \textbf{9.28}\% and \textbf{5.29}\% when 16 bits, 32 bits, 48 bits and 64 bits hash codes are used, compared with CMH. Similar results can be observed in other experiments.
	\item Compared with deep and multi-view methods, for example, D-MVE-Hash achieves gains of \textbf{5.21}\%, \textbf{4.6}\%, \textbf{3.57}\% and \textbf{4.30}\% when 16 bits, 32 bits, 48 bits and 64 bits hash codes are used for retrieval, compared with DCH.
\end{itemize}
As shown in Fig.~\ref{graph:doubleradiu}, when code length increases, similar objects fall into larger hamming radius. On one hand, the hash codes without enhancement are unable to cover the neighborhood of long codes, which causes the worse performance. On the other hand, our method is able to utilize multi-view information, clustering similar objects to smaller hamming radius. As a result, the performance of our method does not worsen when the retrieval hamming radius remains unchanged. In addition to that, the hash codes without enhancement also perform better when the code is short. Take the 32-bit code as an example. As shown in Fig.~\ref{graph:len}, we use fixed length (16 bits) code as basic code to represent the binary code generated from non-multi-view features and the remaining length (16 bits in this case) to represent the binary code generated from multi-view features. Therefore, when the total code length becomes shorter, the code for multi-view is shorter and the misclassification increases, which harms the performance of the part of fixed length basic code. As a result, equal length (e.g., 32 bits) basic code without enhancement could outperform code with multi-view features. To get a full view of the model's performance, we plot the receiver operating characteristic curves for different bits in Fig.~\ref{graph:roc}. Moreover, Fig.~\ref{graph:lossfusion} shows the change of the training loss with the increase of iterations. It turns out that our model can converge to a stable optimum point after about 50,000 training iterations. Fig.~\ref{graph:visual} is the visualization which shows that our D-MVE-Hash produces more relevant results.

\section{Conclusion}

In this paper, we present an efficient approach D-MVE-Hash to exploit multi-view information and the relation of views for hash code generation. The sub-module MV-Hash is a multi-view hash network which calculates view-relation matrix according to the stability of objects in different views. In our framework, we use three enhancement methods to merge view-relation matrix and variety of binary codes learned from single/multi-view spaces into the backbone network. Control experiments indicate that fusion methods has significant contribution to the proposed framework. In addition, we design the memory network to avoid excessive computing resources on view stability evaluation during retrieval. Experiment results and visualization results have demonstrated the effectiveness of D-MVE-Hash and MV-Hash on the tasks of image retrieval.

\ifCLASSOPTIONcaptionsoff
  \newpage
\fi



\bibliographystyle{IEEEtran}
\bibliography{IEEEabrv,paper}
\end{document}